# Real-Time Inference with Large-Scale Temporal Bayes Nets

Masami Takikawa　　Bruce D'Ambrosio
Information Extraction and Transport, Inc.
1600 SW Western Blvd., Suite 300
Corvallis, OR 97333
takikawa@iet.com　　dambrosi@iet.com

Ed Wright
Information Extraction and Transport, Inc.
1911 N. Ft Myer Dr., Suite 600
Arlington, VA 22209
ewright@iet.com

## Abstract

An increasing number of applications require real-time reasoning under uncertainty with streaming input. The temporal (dynamic) Bayes net formalism provides a powerful representational framework for such applications. However, existing exact inference algorithms for dynamic Bayes nets do not scale to the size of models required for real world applications which often contain hundreds or even thousands of variables for each time slice. In addition, existing algorithms were not developed with real-time processing in mind. We have developed a new computational approach to support real-time exact inference in large temporal Bayes nets. Our approach tackles scalability by recognizing that the complexity of the inference depends on the number of interface nodes between time slices and by exploiting the distinction between static and dynamic nodes in order to reduce the number of interface nodes and to factorize their joint probability distribution. We approach the real-time issue by organizing temporal Bayes nets into static representations, and then using the symbolic probabilistic inference algorithm to derive analytic expressions for the static representations. The parts of these expressions that do not change at each time step are pre-computed. The remaining parts are compiled into efficient procedural code so that the memory and CPU resources required by the inference are small and fixed.

## 1 Introduction

Our focus task is real-time situation assessment. Given an input stream of reports from multiple sensors of multiple types in real-time, the task is to infer the current situation, including the entities present, their states, relationships, and activities, quickly enough so that the decision maker can make an immediate intervention if necessary.

Requirements for real-time situation assessment exist in many problem domains. Our experience has been focused on the assessment of cyber attacks on computer networks and in ballistic missile defense. In cyber attack assessment, for example, given an input stream of reports from network intrusion detection systems and intelligence sources, the task is to infer in real-time whether we are currently under cyber attack, by whom, and with what intent. In ballistic missile defense applications, we are interested in the rapid discrimination of warheads from decoys and other objects based on real-time sensor data from multiple sensors. The results of the inference must be available immediately in order for commanders to make engagement decisions.

These tasks have severe real-time constraints. By real-time we mean two things. First, the inference must be efficient. Applications are often required to handle tens or even hundreds of sensor reports per second. Second, the inference should be resource-bounded. By resource, we mean both CPU time and memory use. Often the use of garbage collection or even the "malloc" library call cannot guarantee bounds on CPU-time, so we have developed techniques that avoid memory allocation during real-time inference.

Situation assessment for realistic problems requires a large Bayes net to model multiple entities of interest (e.g., attackers, LANs, and machines), as well as multiple sensors of multiple types (e.g., various kinds of intrusion detection systems). Such a situation may result in hundreds or even thousands of Bayes net nodes for each time slice.

When modelers build these kinds of temporal Bayes nets (TBNs) using standard Bayes net tools, they often mark a subset of nodes as dynamic nodes, index them by time step, and create copies of them for as many time steps as necessary. They keep the number of dynamic nodes as small as possible to make inference tractable. Even so, this method of constructing Bayes nets works only for a small number of time steps.

In order to handle an indefinite number of time steps, a rollup operation is used. New nodes are created for each new time step and old nodes are deleted from the previous time step by absorption, so that the size of the Bayes net remains constant. However, this approach is too computationally intensive to be used in real-time systems. In addition, this approach cannot guarantee time bounds due to the use of memory allocations.

The goal of our research, presented here, is to create a new exact computational approach that makes real-time inference tractable for the large-scale temporal models described above.

This paper is organized as follows. In Section 2, we define a temporal Bayes net that contains both static and dynamic nodes. In Section 3, we describe a fixed computational structure for a TBN. Section 4 includes a discussion of our inference technique. An optimization technique to reduce the complexity of inference is shown in Section 5. We show the feasibility of our approach by a case study in Section 6. Section 7 describes extensions of our approach. Related work is discussed in Section 8. Finally, a concluding remark is given in Section 9.

## 2 Temporal Bayes Nets

In this section, we define a temporal Bayes net (TBN) that contains both static and dynamic nodes.

A TBN is a specialization of a Bayes net (BN). A BN is a directed acyclic graph, in which a node represents a random variable and a directed arc represents a direct influence. Each node possesses a conditional probability table (CPT) that quantifies the effects of the parents on the node. We use the function $\pi(n)$ to represent the set of parents of a node $n$, and $\phi(n)$ to represent its CPT. In this paper, we assume that all variables are discrete.

A BN is constructed by exploiting a conditional independence property that states that a variable is conditionally independent of its non-descendents given its parents. The full joint probability distribution over all variables in a BN can be obtained by multiplying all CPTs in the BN.

**Definition 2.1** A *static node* is a Bayes net node whose value remains constant over the run. We use $S$ to denote the set of all static nodes.

Although it is constant, the value of a static node is uncertain and its belief may change over the run.

**Definition 2.2** A *dynamic node* is a Bayes net node whose value changes over time. It is indexed by time. We use $D_{0:t}$ to denote the set of all dynamic nodes up to time $t$, and $D_t$ for the set of all dynamic nodes at time $t$.

We make two assumptions regarding the dynamic properties of the problem we would like to represent by a TBN.

First, we assume that the current state is conditionally independent of all earlier states given the immediately preceding state. Second, we assume that each dynamic node has the same CPT regardless of time.

We define *transitional nodes* to formalize these assumptions:

**Definition 2.3** We divide $D_t$ into two sets. Nodes from which arcs originate and are directed into the next time step are called *transitional* nodes. All other dynamic nodes are referred to as *non-transitional* nodes. We denote them by $T_t$ and $N_t$, respectively.

**Assumption 2.1** We assume that a dynamic node is not a parent of a static node, that is, $\forall s \in S : \pi(s) \subset S$. In addition, we assume that each parent of a node in $D_i$ is either a static node, a dynamic node at the same time step, or a transitional node at the previous time step, that is, $\forall i : \forall d \in D_i : \pi(d) \subset S \cup D_i \cup T_{i-1}$.

**Assumption 2.2** For each dynamic node $d \in D_0$, the CPTs of $d_i$ for all $i >= 0$ are the same under appropriate renaming of variables, that is, $\forall d \in D_0 : \forall i \geq 0 : \phi(d_i) \cong \phi(d)$.

For example, if the parents of $d_5$ are $[s, n_5, t_4]$, where $s$ is a static node, $n_5$ is a non-transitional node, and $t_4$ is a transitional node, then under this assumption the parents of $d_6$ are $[s, n_6, t_5]$, and the numbers in the CPT for $d_5$ match those in the CPT for $d_6$.

To complete a Bayes net, we need a set of initial transitional nodes $T_{-1}$ and their initial CPTs.

**Definition 2.4** A *fully-extended temporal Bayes net for t* (FE-TBN$_t$) is a Bayes net that consists of a set of nodes that are equal to the union of $S$, $N_{0:t}$, and $T_{-1:t}$, a set of directed acyclic arcs among them that satisfy Assumption 2.1, and a set of CPTs that satisfy Assumption 2.2.

As with the standard Bayes nets, the joint probability distribution over all nodes in a FE-TBN$_t$ can be obtained by multiplying all of its CPTs, so the FE-TBN$_t$ can answer any temporal probabilistic query up to time $t$.

In this paper, we are interested in a query that asks the probability of a static node given evidence on dynamic nodes. For example, what are the characteristics of the attacker given the intrusion detection reports?

For convenience, we denote the set of observable dynamic nodes by $O_{0:t}$. We assume that for each observable node, the function $\lambda$ returns a likelihood vector[1] representing the observation, so that the posterior full joint probability distribution may be obtained by the multiplication of $\lambda$ with the prior full joint probability distribution and normalization of the result. Also, we denote the target static variable by $s$. The query of interest can now be expressed as $P(s|\lambda_{0:t})$. (See Figure 1.)

---

[1] Note that if there is no observation for a particular observable, $\lambda$ returns a uniform likelihood vector.

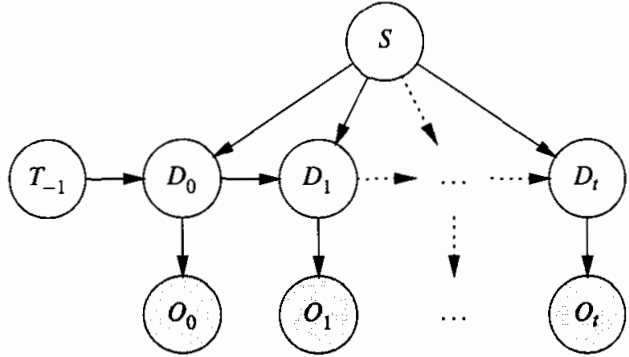

Figure 1: FE-TBN$_t$

We can compute this query by the following equation:

$$P(s|\lambda_{0:t}) = \alpha \sum_{V \setminus \{s\}} \Phi(S)\Phi(T_{-1}) \prod_{i=0}^{t} \Phi(D_i)\Lambda(O_i) \quad (1)$$

where $\alpha$ is a normalization constant, $V = S \cup N_{0:t} \cup T_{-1:t}$ is the set of all nodes, $\Phi(V) = \prod_{v \in V} \phi(v)$ is the product of the CPTs of the given set of nodes, and $\Lambda(O) = \prod_{o \in O} \lambda(o)$ is the product of the likelihood vectors of the given set of observable nodes.

Although a FE-TBN$_t$ can answer any query up to time $t$, it grows in size as $t$ increases. For this reason, it cannot be used for an indefinite number of time steps. Our goal is to develop a computational approach that can answer in real-time the query $P(s|\lambda_{0:t})$ for any $t$ while avoiding memory increase.

## 3 Fixed Temporal Bayes Nets

In this section, we describe our computational approach in which a fixed computational structure is used to answer the query $P(s|\lambda_{0:t})$.

Figure 2 shows a two-time-slice part of a small example TBN where the nodes $a$, $b$, and $c$ are static, $d$ is a non-transitional dynamic node, $e$ is a transitional dynamic node, and $f$ is an observable non-transitional dynamic node. As can be seen, for each time step, the subgraph of all dynamic nodes has the same internal structure and external interface, along with the same parameters due to Assumption 2.2.

**Definition 3.1** The *static parents*, denoted by $R$, are the set of static nodes, each of which is a parent of a dynamic node, that is, $R = S \cap \{p | p \in \pi(d) \text{ for } d \in D_{0:t}\}$.

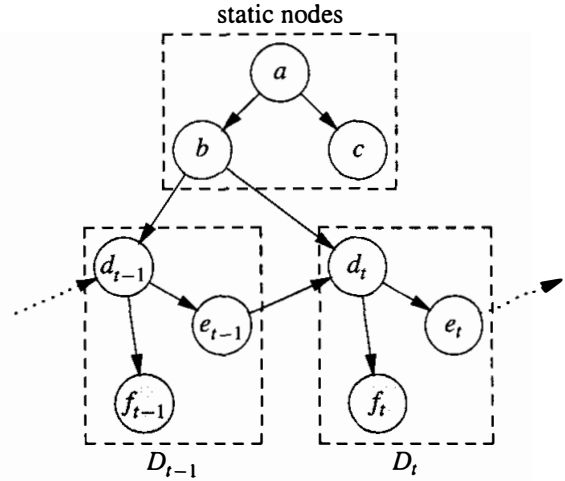

Figure 2: Partial TBN for two time slices

**Definition 3.2** The *interface nodes* at time $t$, denoted by $I_t$, are the union of static parents and transitional nodes at $t$, that is, $I_t = R \cup T_t$.

For example, in Figure 2, $R = \{b\}$ and $I_t = \{b, e_t\}$.

**Definition 3.3** The *past expression*, denoted by $\psi(I_t)$, is a multidimensional array over the interface nodes at time $t$, and is defined as follows:

$$\psi(I_t) = \beta \sum_{N_{0:t} \cup T_{-1:t-1}} \Phi(T_{-1}) \prod_{i=0}^{t} \Phi(D_i)\Lambda(O_i) \quad (2)$$

This equation multiplies the CPTs from all dynamic nodes and observations, and marginalizes out all dynamic nodes but the current transitional nodes. $\beta$ is a normalization constant that normalizes the entire table so that the total probability mass of the table becomes one.[2]

The past expression summarizes all observations and is the only information that is carried from the current time step to the next time step.

This equation can be rewritten using $\psi$ recursively as follows, which is the basis of our computational structure:

$$\psi(I_t) = \beta \sum_{N_t \cup T_{t-1}} \psi(I_{t-1})\Phi(D_t)\Lambda(O_t) \quad \text{for } t \geq 0$$
$$\psi(I_{-1}) = \Phi(T_{-1})$$
$$(3)$$

The target query in Equation 1 can be written using the past expression as follows:

---
[2]Note that normalization is not mathematically necessary here, but is required to avoid underflow in the floating-point calculation.

$$P(s|\lambda_{0:t}) = \alpha \sum_{S \cup T_t \setminus \{s\}} \Phi(S)\psi(I_t) \quad (4)$$

Now we define the fixed TBN which exploits the above equations to create a fixed Bayes net:

**Definition 3.4** The *fixed TBN* (FX-TBN) for a FE-TBN$_t$ is a Bayes net that consists of a set of nodes, $S \cup D_t \cup T_{t-1}'$. The arcs and CPTs are the same as those of the underlying FE-TBN$_t$, except those for $T_{t-1}'$, whose CPTs are collectively defined as $\psi(I_{t-1})$.

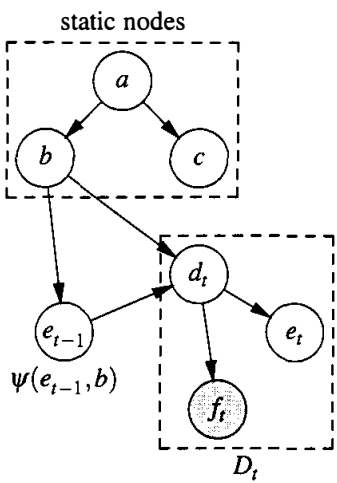

Figure 3: Example Fixed-TBN

Figure 3 shows a FX-TBN created from the example shown in Figure 2.

FX-TBNs differ from standard Bayes nets in the definition of CPTs for the nodes in $T_{t-1}'$. In FX-TBNs, the single CPT, $\psi(I_{t-1})$, is used for all nodes in $T_{t-1}'$. Recall that in standard Bayes nets, each node possesses its own CPT. However, any BN inference algorithm can be easily adapted to handle FX-TBNs.

Using a FX-TBN, the rollup operation becomes the following simple procedure, which we call the *advance* procedure:

1. Compute $\psi(I_{-1})$ and make it the CPT of $T_{-1}'$.
2. Make observations for $O_t$.
3. Query any node of interest.
4. Compute $\psi(I_t)$ and make it the CPT of $T_{t-1}'$.
5. Repeat from Step 2.

For each iteration, the time step increases by one. The correctness of this procedure can be stated as follows:

**Theorem 3.1 Correctness** After $t$ iterations of the above advance procedure, a FX-TBN represents the same probability distribution over its nodes as the corresponding FE-TBN$_t$.

This theorem can be proven by Equation 3 and Assumption 2.2.

The complexity of this procedure depends on the underlying Bayes net structure, and is at least the size of the past expression, which is exponential in the number of interface nodes (i.e., the union of transitional nodes and static parents). Thus, the applicability of our approach is determined primarily by the size of the interface for a particular application. Non-interface nodes (i.e., non-transitional dynamic nodes and static nodes whose children don't include a dynamic node) are irrelevant to inference complexity with respect to the past expression.

## 4 Inference with Fixed TBNs

Although any BN inference algorithm can be used with a FX-TBN, most existing algorithms are not suitable for real-time processing. In this section, we describe our technique for compiling a FX-TBN query and the advance procedure into efficient procedural code with fixed resource requirements.

We compute two expressions, namely, the past expression defined in Equation 3 and the target query defined in Equation 4. Both equations have the same abstract structure, that is, a product of multiple expressions followed by a marginalization. Their computational cost can be reduced when nodes are summed early in the computation, rather than performing all marginalization after computing the full product.

Given the expressions to be multiplied, we can create a binary tree in which leaves are the input expressions and each node represents the product of two children. We call this tree a *factoring tree*. As soon as all expressions that contain a node $x$ are multiplied, $x$ can be marginalized out. Thus, the core of any BN inference algorithm can be seen as the task of finding an optimal factoring tree that has the smallest cost (e.g., the total number of floating-point multiplications).[3] This task has been proven to be NP-hard, but simple greedy heuristics work well in most cases.

---

[3]Note that both the junction-tree and variable-elimination algorithms have been proven to be suboptimal, i.e., there exists a BN for which those algorithms cannot find an optimal factoring tree. SPI is the only algorithm to our knowledge that is potentially optimal. For example, consider a net consisting of node $h$ with parents $a$, $b$, $c$, and $d$, and a marginal query for $h$. Li and D'Ambrosio [1994] showed that an optimal sequence is to combine $\phi(a)$ with $\phi(b)$, and $\phi(c)$ with $\phi(d)$, and then to combine either with $\phi(h)$. However, this sequence cannot be generated from the variable-elimination perspective, since no matter which variable we choose to eliminate first, the algorithm will immedi-

Given expressions and observables, our algorithm works as follows:

1. Utilizing d-separation, find the minimum set of expressions needed to compute the past expression and the target query.

2. Find the best factoring trees possible for them using heuristics.

3. Prune branches of the trees that do not contain observables. The results of these branches do not change over time, so they are reused at each time step.

4. Compile the remaining branches into procedural code.

The resulting code contains three operations: posting observations to observables, computing the target query, and advancing the time step by updating the past expression. It uses only primitive machine operations, such as multiplying floating-point numbers and accessing memory, and does not use any library calls, including memory allocation. The product of two expressions is computed by a nested for-loop of depth equal to the number of variables involved. Thus, the resulting code is small and extremely efficient, and its memory and CPU requirements are fixed.

## 5 Reducing Inference Complexity

In this section, we talk about a technique to reduce inference complexity by exploiting the distinction between dynamic and static nodes.

As we discussed in Section 3, the past expression contains all interface nodes, so its size is exponential in the number of those nodes. When the interface nodes are independent, we can reduce the size of the past expression by factorizing it into smaller tables. Although they are usually not independent, they are often conditionally independent given static nodes.

Let us show the impact of this optimization technique using the example shown in Figure 4.

In this figure, the node $a$ is static and the remaining nodes are dynamic. Nodes $b$, $c$, and $d$ are transitional. The interface nodes are $a$, $b$, $c$, and $d$. The three transitional nodes are not independent, but they are conditionally independent given $a$. Thus, we can factorize the past expression as follows:

$\psi(b_t, c_t, d_t, a) = \psi(b_t, a)\psi(c_t, a)\psi(d_t, a)$
where
$\psi(b_t, a) = \beta \sum_{\{e_t, b_{t-1}\}} \psi(b_{t-1}, a)\phi(b_t)\phi(e_t)\lambda(e_t)$
$\psi(c_t, a) = \beta \sum_{\{f_t, c_{t-1}\}} \psi(c_{t-1}, a)\phi(c_t)\phi(f_t)\lambda(f_t)$
$\psi(d_t, a) = \beta \sum_{\{g_t, d_{t-1}\}} \psi(d_{t-1}, a)\phi(d_t)\phi(g_t)\lambda(g_t)$

ately combine the CPT for that node with $\phi(h)$. Thus, it fails to find the optimal factoring.

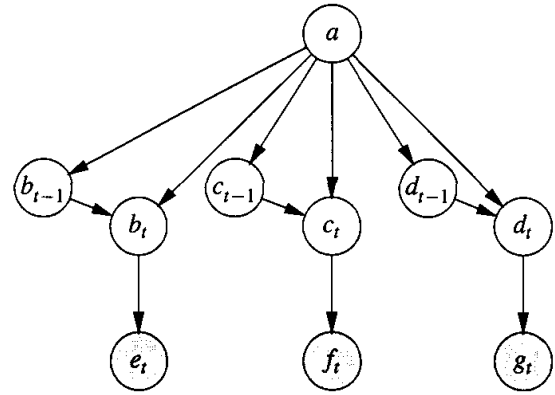

Figure 4: A TBN with three independent transitions given the static node

This kind of factorization can reduce the complexity of inference significantly and is applicable to many domains. For example, in a cyber attack situation assessment, we can model the status of each computer network host with a transitional node as in Figure 4, and the current defense condition of the national computer network as a static node. With this representation, we can perform inference with a large number of hosts without creating a prohibitively huge past expression.

The factorization of a past expression can be found as a byproduct of computing the past expression. Let's suppose that we use the variable elimination (VE) algorithm [Zhang and Poole, 1996], which marginalizes out nodes, one by one, by multiplying all expressions that involve the variable to be eliminated. When computing a past expression using VE given the factorized past expression of the previous step, after eliminating all non-interface nodes, there remains a set of expressions left to be combined. This set of expressions is the factorization of the past expression we are looking for. Because the resulting factorization is the same regardless of the elimination ordering, and because we are only interested in the symbolic structure of the result, we can compute the factorization in $O(n^3)$ where $n$ is the number of nodes.[4]

This method of computing the factorization of the past expression poses a tricky problem to the query compiler. As discussed in the previous section, the query compiler requires the same factoring tree for all time steps, but the factorization of the past expression depends on that of the previous past expression, and may vary. For example, consider the TBN shown in Figure 5, where $a$ is static, $b$, $c$, and $d$ are transitional, and $e$, $f$, and $g$ are observable.

In this TBN, the factorization of the past expression af-ter the first advance procedure is $\psi(b_t,a)\psi(c_t)\psi(d_t)$. The next application of the advance procedure changes this to $\psi(b_t,a)\psi(c_t,a)\psi(d_t)$. Finally, after the third iteration, the factorization becomes $\psi(b_t,a)\psi(c_t,a)\psi(d_t,a)$, and we have the following stable computational structure:

$$\psi(b_t,c_t,d_t,a) = \psi(b_t,a)\psi(c_t,a)\psi(d_t,a)$$
where
$$\psi(b_t,a) = \beta \Sigma_{\{e_t,d_{t-1}\}} \psi(d_{t-1},a)\phi(b_t)\phi(e_t)\lambda(e_t)$$
$$\psi(c_t,a) = \beta \Sigma_{\{f_t,b_{t-1}\}} \psi(b_{t-1},a)\phi(c_t)\phi(f_t)\lambda(f_t)$$
$$\psi(d_t,a) = \beta \Sigma_{\{g_t,c_{t-1}\}} \psi(c_{t-1},a)\phi(d_t)\phi(g_t)\lambda(g_t)$$

---

[4]$O(n^3)$ follows from the fact that there are at most $n$ variables to eliminate, each variable belongs to at most $n$ expressions, and each expression contains at most $n$ variables.

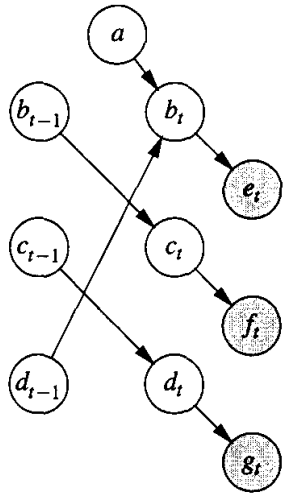

Figure 5: A TBN with varying factorization

As can be seen from this example, the factorization stabilizes after at most $n$ iterations, where $n$ is the number of transitional nodes.

## 6 Case Study

In this section, we show the experimental results obtained from a case study.

We used the model developed for the SSARE (Security Situation Assessment and Response Evaluation) project [D'Ambrosio et al., 2001] in the DARPA Information Assurance program. The SSARE model was developed using JPF [D'Ambrosio et al., 2000], an object-oriented probabilistic knowledge representation language. Using JPF, the modeler describes generic probabilistic knowledge about entities and their relations using frames and creates a situation-specific Bayes net model by instantiating the frames and connecting those instances. JPF uses local expression languages [D'Ambrosio, 1995] to efficiently factorize the noisy-max [Takikawa and D'Ambrosio, 1999] and the multiplexor (i.e., $P(x|s,x_1,x_2,\ldots,x_n) = P(x|x_i)$ if $s = i$), both of which are heavily used by the SSARE model. JPF employs symbolic probabilistic inference (SPI) [Shachter et al., 1990] to perform BN inference.

Given attack reports from intrusion detectors and status reports from host machines, the SSARE model tries to answer many questions including the following: Are we under attack? Who is attacking? What kind of attacks are we experiencing?

The original model was developed with limited support for time. We modified the model by representing the status of each machine by a transitional dynamic node. The BN generated by JPF is a fixed TBN. Table 1 shows some metrics of this Bayes net.

| Total Number of Nodes | 792 |
|---|---|
| Number of Static Nodes | 472 |
| Number of Dynamic Nodes | 320 |
| Number of Transition Nodes | 23 |
| Number of Static Parents | 13 |
| Number of Dynamic Observables | 9 |

Table 1: Metrics of SSARE model

Our target query for this experiment was to determine whether or not a particular agent was attacking. This query involved 413 nodes and required 2,864,345 multiplications, with the largest intermediate table of size 419,904.

The past expression was factorized into 15 expressions. The largest factor contained 15 nodes and 1,259,712 numbers. Computing the past expression involved 155 nodes and required 18,225,110 multiplications with the largest intermediate table of size 1,259,712.

In this experiment, we used C++ as the target procedural language. The C++ code, which was compiled from the Bayes net to compute the target query and the past expression, contained a constant input table of size 1,313,926 of which 53,587 entries corresponded to the precomputed branch. It used a table of size 20,818,246 to store intermediate and final results. These tables were statically allocated at the start and no memory allocation occurred during the run.

The C++ code was able to perform two advance operations per second using an Intel Pentium 4 CPU 1.5GHz.

## 7 Extensions

### 7.1 Hierarchical TBNs

What are static nodes? They are the nodes that stay the same through the run (although they are uncertain and their belief changes over time). Although one can consider, say, the computer network topology to be static, it certainly will change in the long run. Our computational approach can easily be extended to handle such cases.

Consider the hierarchy of time durations as a tree. The root spans the entire time. Nodes that are truly static nodes belong in the root. The next level divides time into large intervals. The nodes that change their values slowly belong in this level. The third level divides the time intervals further, and nodes in this level mutate faster. We can continue creating more levels as necessary.

Now, let's assume that for all $i$, no node at level $i$ has a node at level greater than $i$ as a parent. Then, we can apply our computational approach to advance each level $i$ by considering all nodes at $i$ or below as dynamic nodes, and all nodes at $i-1$ or above as static nodes.

### 7.2 Multiple Time Steps

Although a FX-TBN contains only one complete time step, extending it to contain multiple time steps can be easily done without sacrificing the ability to perform real-time inference with a fixed computational structure.

There are two notable applications of this extension. First, it can be used to solve sequential decision problems by encoding a finite-horizon approximation of the problem.

Second, it can be used to accommodate evidence that does not arrive in temporal order. Note that in this paper we assume that evidence arrives in temporal order. This assumption may seem obvious, but evidence that arrives out of order is a concern to the Defense Data Fusion community. For example, HUMINT (intelligence derived from human sources) can arrive very slowly often more slowly than information from other sources. Also, with distributed sensors communication delays may further contribute to out-of-order evidence arrival. One possibility to handle such cases is to retain multiple time steps so that we can post late-arriving observations to some past observable.

## 8 Related Work

Dynamic Bayes nets (DBN) [Russell and Norvig, 1995] can be seen as a specialization of the temporal Bayes nets discussed in this paper (i.e., in a DBN all nodes are dynamic). Because static nodes are represented by dynamic nodes whose transition matrices are the identity in the DBN, static nodes are included in the interface nodes. This will make it intractable to apply the standard DBN approach to the kind of networks we are dealing with here.

D'Ambrosio [1992] describes an online maintenance agent that uses dynamic influence diagrams similar to DBNs to perform real-time diagnosis. It moves in time by a rollup operation, adding new nodes for each subsequent time step, and deleting old nodes from the previous time step. The inference is done by *term computation*, which is an anytime approximate inference algorithm. In order for term computation to be effective, the Bayes net must be *skewed* so that most of probability mass is contained in the largest few terms. In contrast, our approach uses an exact inference algorithm and does not make any assumptions about the probability distribution.

DHugin [Kjærulff, 1995] applies the junction-tree algorithm to DBNs. It also uses a rollup operation, but avoids creating a junction-tree from scratch at each time step by constraining the elimination ordering so that older nodes are eliminated before newer ones, and by modifying the junction-tree incrementally. The use of the junction-tree increases the complexity of inference compared to our approach because the interface nodes must contain the parents of nodes with incoming temporal arcs in order to create the moral graph. In our approach, those nodes are not necessarily among the interface nodes. The incremental modification of the junction-tree makes it difficult to apply DHugin to real-time systems.

Recently, we found that Murphy [2001] uses a single junction-tree to process an indefinite number of time steps, avoiding an expensive rollup operation. Our approach can be seen as an extension of his approach to temporal Bayes nets with static nodes, so as to reduce the size of interface nodes and factorize the interface. The increase in the size of the interface of the junction-tree described above also applies to this approach.

The idea that, conditioned on some variables, inference on the remaining substructures can be performed efficiently, has been used in both static and dynamic Bayes nets. For example, Pearl's cutset conditioning [1988] is an application of this idea to static Bayes nets. Murphy and Russell [2000] apply this idea to approximate inference in dynamic Bayes nets by conditioning some variables by sampling.

Friedman *et al.* [1998] present a structured representation system that combines DBNs with object-oriented Bayes nets. It can represent complex stochastic systems similar to the TBNs we describe in this paper. We believe that our computational approach is also applicable to their representation.

Some research systems have been developed that generate efficient executable code from the probabilistic models. For example, AutoBayes [Fischer and Schumann, 2002] generates data analysis programs from statistical models, and CES [Thrun, 1998] is an extension of the programming language C, augmented with various constructs to manipulate probability distributions. Darwiche and Provan [1997] describe a system that is closely related to the code generator of our system. Their system generates a DAG from a Bayes net query. A non-leaf node in the DAG represents a number, evidence, or a numeric operation. Each leaf node corresponds to an answer to the given query. The query DAGs are finer-grained than the factoring trees in our system, in which a node represents an expression, so they support finer optimization of the computation, but at the cost of a constant-factor increase in memory. Another difference is that the DAGs are interpreted by a compact interpreter, while our system generates compact machine code that can be executed directly by machines.

## 9 Conclusions

We have presented a new computational approach to real-time inference with large-scale temporal Bayes nets and shown its feasibility by a case study in which we were able to perform multiple advance operations per second with a fairly large temporal Bayes net.

As shown in our case study, the combination of a high-level probabilistic representational framework and an efficient computational approach is very powerful. We believe that our work is an important step toward developing an optimizing compiler for a high level probabilistic language that generates procedural code that is more efficient than humans can write. This capability will allow modelers to rapidly develop models of complex problem domains and then automatically generate efficient executable code.

**Acknowledgements**

The ideas in this paper have benefited from discussion with Suzanne Mahoney and Dan Upper of IET, and members of MDA Hercules Program. We thank Jane Jorgensen, Francis Fung, and Kathy Laskey of IET, and the anonymous reviewers for valuable comments on drafts of this paper. Funding for this research is provided in part under the MDA (formerly, BMDO) Hercules Program contract # HQ0006-02-C-0004.

**References**

[D'Ambrosio et al., 2000] B. D'Ambrosio, M. Takikawa, and D. Upper. Representation for dynamic situation modeling. Technical report, Information Extraction and Transport, Inc., 2000.

[D'Ambrosio et al., 2001] B. D'Ambrosio, M. Takikawa, J. Fitzgerald, D. Upper, and S. Mahoney. Security situation assessment and response evaluation (SSARE). In *Proceedings of the DARPA Information Survivability Conference & Exposition II*, volume I, pages 387–394. IEEE Computer Society, 2001.

[D'Ambrosio, 1992] B. D'Ambrosio. Real-time value-driven diagnosis. In *Proceedings, Third International Workshop on the Principles of Diagnosis*, 1992.

[D'Ambrosio, 1995] B. D'Ambrosio. Local expression languages for probabilistic dependence. *International Journal of Approximate Reasoning*, 13:61–81, 1995.

[Darwiche and Provan, 1997] A. Darwiche and G. Provan. Query dags: A practical paradigm for implementing belief-network inference. *Journal of Artificial Intelligence Research*, 6:147–176, 1997.

[Fischer and Schumann, 2002] B. Fischer and J. Schumann. Generating data analysis programs from statistical models. *Journal of Functional Programming*, 2002. to appear.

[Friedman et al., 1998] N. Friedman, D. Koller, and A. Pfeffer. Structured representation of complex stochastic systems. In *Proceedings of the Fifteenth National Conference on Artificial Intelligence*, pages 157–164, 1998.

[Kjærulff, 1995] U. Kjærulff. dHugin: A computational system for dynamic time-sliced bayesian networks. *International Journal of Forecasting*, 11:89–111, 1995.

[Li and D'Ambrosio, 1994] Z. Li and B. D'Ambrosio. Efficient inference in Bayes networks as a combinatorial optimization problem. *International Journal of Approximate Reasoning*, 11:1–24, 1994.

[Murphy and Russell, 2000] K. Murphy and S. Russell. Rao-Blackwellised particle filtering for dynamic bayesian networks. In A. Doucet and N. de Freitas, editors, *Sequential Monte Carlo Methods in Practice*. 2000.

[Murphy, 2001] K. Murphy. Applying the junction tree algorithm to variable-length DBNs. Technical report, Computer Science Department, University of California, Berkeley, 2001.

[Pearl, 1988] J. Pearl. *Probabilistic Reasoning in Intelligent Systems: Networks of Plausible Inference*. Morgan Kaufmann, 1988.

[Russell and Norvig, 1995] S. Russell and P. Norvig. *Artificial Intelligence: A Modern Approach*. Prentice Hall Series in Artificial Intelligence. Prentice-Hall Inc., 1995.

[Shachter et al., 1990] R. Shachter, B. D'Ambrosio, and B. DelFavero. Symbolic probabilistic inference in belief networks. In *Proceedings Eighth National Conference on AI*, pages 126–131. AAAI, 1990.

[Takikawa and D'Ambrosio, 1999] M. Takikawa and B. D'Ambrosio. Multiplicative factorization of noisy-max. In *Proceedings of the Fifteenth Conference on Uncertainty in Artificial Intelligence*, pages 622–630, 1999.

[Thrun, 1998] S. Thrun. A framework for programming embedded systems: Initial design and results. Technical Report CMU-CS-98-142, Department of Computer Science, Carnegie-Mellon University, 1998.

[Zhang and Poole, 1996] N. L. Zhang and D. Poole. Exploiting causal independence in Bayesian network inference. *Journal of Artificial Intelligence Research*, 5:301–328, 1996.